\documentclass[sigconf]{acmart}
\usepackage{subfig}

\AtBeginDocument{%
  \providecommand\BibTeX{{%
    \normalfont B\kern-0.5em{\scshape i\kern-0.25em b}\kern-0.8em\TeX}}}

\copyrightyear{2021}
\acmYear{2021}
\setcopyright{acmcopyright}
\acmConference[MMSports '21] {Proceedings of the 4th International Workshop on Multimedia Content Analysis in Sports}{October 20, 2021}{Virtual Event, China.}
\acmBooktitle{Proceedings of the 4th International Workshop on Multimedia Content Analysis in Sports (MMSports '21), October 20, 2021, Virtual Event, China}
\acmPrice{15.00}
\acmISBN{978-1-4503-8670-8/21/10}
\acmDOI{10.1145/XXXXXX.XXXXXX}

\begin{document}

\title {Multi-task learning for jersey number recognition in Ice Hockey}

\author{Kanav Vats, Mehrnaz Fani, David A. Clausi and John Zelek}
\email{{k2vats,mfani,dclausi,jzelek}@uwaterloo.ca}
\affiliation{%
  \institution{University of Waterloo}
  \streetaddress{}
  \city{Waterloo}
  \state{Ontario}
  \country{Canada}
  \postcode{}
}

\begin{abstract}
Identifying players in sports videos by recognizing their jersey numbers is a challenging task in computer vision. We have designed and implemented a multi-task learning network for jersey number recognition. In order to train a network to recognize jersey numbers, two output label representations are used (1) Holistic - considers the entire jersey number as one class, and (2) Digit-wise - considers the two digits in a jersey number as two separate classes. The proposed network learns both holistic and digit-wise representations through a multi-task loss function. We determine the optimal weights to be assigned to holistic and digit-wise losses through an ablation study. Experimental results demonstrate that the proposed multi-task learning network performs better than the constituent holistic and digit-wise single-task learning networks.
\end{abstract}






\maketitle

\section{Introduction}
Automated player identification in team sports using  broadcast game video is a challenging task. Although there are approaches for some sports (e.g., basketball \cite{Senocak_2018_CVPR_Workshops}) that identify players using body appearances, this is not achievable in ice hockey due to the players wearing bulky equipment and helmets that occlude body characteristics and skin color, especially reducing the discriminability between players on the same team that wear the same color uniforms and helmets.  For ice hockey, this leaves jersey numbers as the primary method of performing player identification from game video.  Although approaches for digit recognition exist \cite{goodfellow}, high motion blur, occlusions, and single camera views make this a challenging automation problem for ice hockey.  

In the literature, there exist several deep learning approaches for jersey number recognition \cite{gerke,li,liu,GERKE2017105}. These approaches consider jersey number recognition as a classification problem and either (1) consider the jersey numbers as separate classes \cite{gerke,GERKE2017105}, or (2) treat the two digits in a jersey number as two independent classes \cite{li,liu}. Since learning multiple output representations through multi-task learning can lead to improved regularization \cite{ruder}, we hypothesize that learning both of these representation together in a multi-task loss can result in better performance.

\begin{figure*}[t]
\begin{center}
\includegraphics[width=\linewidth]{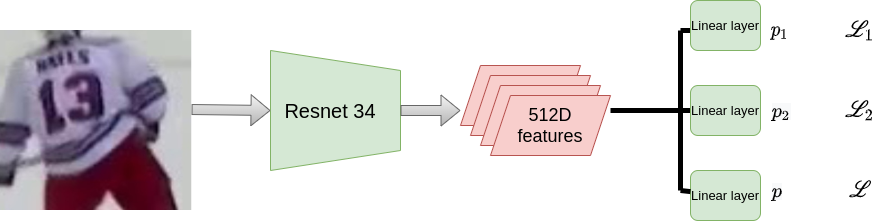}
\end{center}
  \caption{The input image is passed through a Resnet 34\cite{resnet} network after which the $512$ dimensional features are extracted from the pre-final layer. $\{p_i,i \in \{1,2\} \}$ and $p$ are 11 and 81-dimensional vectors representing the digit probabilities and holistic number probabilities respectively.  $\mathcal{L}_1$ and $\mathcal{L}_2$ denotes the individual first and second digit loss respectively and $\mathcal{L}$ denotes the holistic loss. }
\label{fig:probs_game_time}
\end{figure*}

In this paper, we utilize multi-task learning for simultaneously learning the digit-wise and holistic jersey number representations for improving network generalization. Two contributions are recognized: \begin{enumerate}
    \item  We design a loss function consisting of the combination of of (1) "Holistic" representation loss term treating the jersey number as a separate class (2) "Digit-wise" representation loss term treating digits in a number as independent classes.
    \item  Due to lack of publicly available datasets for jersey number recognition, we also introduce a jersey number dataset consisting of more than $50,000$ images for the game of ice hockey.
\end{enumerate}
We  conduct an ablation study to demonstrate that the holistic and digit-wise losses complement each other with appropriate weight given to them. Experimental results demonstrate that the proposed multi-task loss performs better than using only the holistic loss or the digit-wise loss.

\section{Background}

Traditionally, jersey number recognition was performed with the help of hand-crafted features \cite{Saric2008PlayerNL, hc}. Gerke \textit{et al.} \cite{gerke} were the first to use deep networks (CNN) for jersey number recognition with small low resolution images. The CNN outperformed hand crafted HOG features by a huge margin. Li \textit{et al.} \cite{li} use a spatial transformer network (STN) to better localize jersey numbers and use human labeled quadrangle, annotating the jersey number area, to train the network with semi-supervised learning. Liu \textit{et al.} \cite{liu} treat the problem as a jersey number detection and recognition problem using a Faster RCNN \cite{fasterrcnn} inspired network incorporating human pose keypoint supervision. Gerke \textit{et al.} \cite{GERKE2017105} complement vision based jersey number recognition model with player location based features based on the assumption that players do not move randomly on the field, but follow a tactical role such as defender, winger or forward. \\
In the literature, multi-task learning has been utilized for digit and text recognition. How \textit{et al.} \cite{hou} use a multi-task learning network to simultaneously learn handwritten numeral recognition and determine whether the digit is scratchy. Kim \textit{et al.} \cite{kim} use multi-task learning for determining the presence of license plates, recognizing plate digits and plate characters altogether.\par

\section{Methodology}
\subsection{Dataset}

Datasets used in recent works \cite{gerke,liu,li,GERKE2017105} are not publicly available, hence we created our own dataset. The dataset consists of  $54,251$ player bounding boxes obtained from 25 National Hockey League (NHL) games. The NHL game videos are of resolution $1280 \times 720$ pixels. The dataset contains a total of $81$ jersey number classes, including an additional null class for no jersey number visible. The dataset is much bigger than the datasets used in other works such as Gerke \textit{et al.} \cite{gerke} with 8,281 images and Liu \textit{et al.} \cite{liu} with 3,567 images and 6,293 digit instances (Table \ref{table:dataset_comp}). Although the dataset used in Li \textit{et al.} \cite{li} has $215,036$ images, $90\%$ of the images are negative samples (no jersey number present). Hence, our dataset has more images with a non-null jersey number  than Li \textit{et al.} \cite{li}. \par
\begin{figure*}
	\begin{center}
		\subfloat[]{
		    \includegraphics[width=0.25\linewidth]{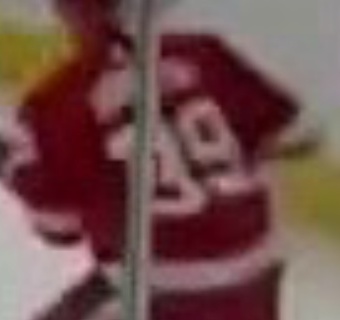}
		}
		\subfloat[]{
		    \includegraphics[width=0.25\linewidth]{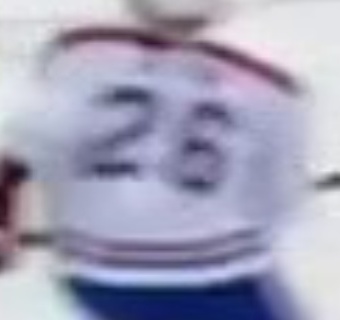}
		}
		\subfloat[]{
		    \includegraphics[width=0.25\linewidth]{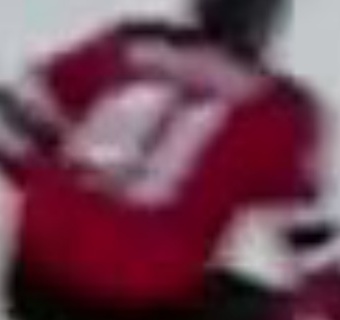}
		}
		\subfloat[]{
		    \includegraphics[width=0.25\linewidth]{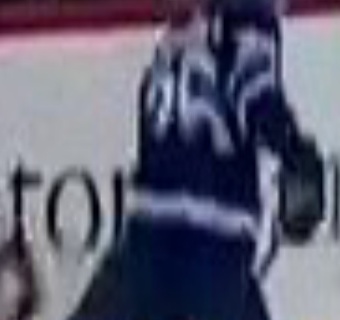}
		}
	\end{center}
	\caption{Examples of images from the dataset. The dataset cover many real-game scenarios such as (a) occlusions from external objects, (b)(c) motion blur, and (d) self-occlusion.}
	\label{fig:images_expl}
\end{figure*}

\begin{figure}[t]
\begin{center}
\includegraphics[width=\linewidth, height = 4cm]{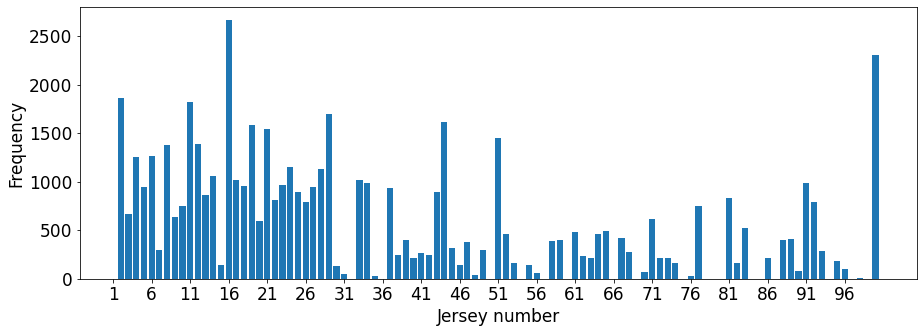}
\end{center}
  \caption{Class example distribution in the dataset. the total number of classes is 81 and this includes the ``not visible" class. The dataset is highly imbalanced such that the ratio between the most frequent and least frequent class is $92$. }
\label{fig:class_dist}
\end{figure}

 
The player head and bottom of the images are cropped such that only the jersey number is visible. Fig. \ref{fig:images_expl} shows some example images from the dataset. A number was considered readable when both constituent digits were visible, however, images with partial occlusion due to motion blur and jersey kinks were included in the dataset since those situations are very common and a model working in sports scenarios should handle those situations.  A digit was considered unreadable when either one/both of its constituent digits was fully occluded/invisible. Two annotators annotated the entire dataset. \\Images from $17$ games are used for training, four games for validation and four games for testing. The exact number of images in the splits is shown in  Table ~\ref{table:split}.  The splits are constructed at a game level, so that there is no inherent in-game bias present during validation or testing. The dataset is highly imbalanced such that the ratio between the most frequent and least frequent class is $92$. The class distribution in the dataset is illustrated in Fig. \ref{fig:class_dist}. The dataset covers a range of real-game scenarios such as occlusions, motion blur and self occlusions. We plan on making the dataset publicly available in future.

\begin{table}[!t]

    \centering
    \caption{Number of images in train, validation and test set }
    \footnotesize
    \setlength{\tabcolsep}{0.2cm}
    \begin{tabular}{c|c|c}\hline
  
         Train & Validation & Test \\\hline
      $38,456$  & $6,770$ & $9,025$ \\
      
    \end{tabular}
    \label{table:split}
\end{table}

\subsection{Network Design}
To solve the previously described problem, i.e., players' jersey number recognition in broadcast ice hockey videos, a network with a multi-task loss, as shown in Fig. \ref{fig:probs_game_time}, is designed and implemented. The input image of dimension $300 \times 300$ pixels is passed through a Resnet34 \cite{resnet} network  to obtained 512-dimensional features from the pre-final layer. The features are then passed through three linear layers followed by softmax layers to output three probabilities. The first linear layer outputs an 81-dimensional vector $p \in R^{81}$ representing the probability distribution over the 81 jersey number classes. The second and third linear layers output an 11-dimensional vectors $p_1, p_2 \in R^{11}$ representing the probability of the first and second digit respectively. The one additional class in the 11-dimensional vector denotes the absence of a jersey number. Let $y \in R^{81}, y_1 \in R^{11}$ and $y_2 \in R^{11}$ denote the ground truth vectors corresponding to the jersey number, first digit and second digit respectively. \\

\begin{figure}[t]
\begin{center}
\includegraphics[width=8.5cm, height=5.5cm]{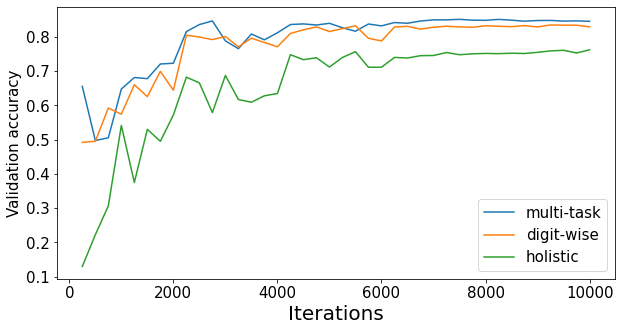}
\end{center}
  \caption{Validation accuracy vs number of iterations for the multi-task learning(MTL), holistic and digit-wise loss settings. The multi-task setting shows the best performance among the three settings. }
\label{fig:val_acc}
\end{figure}
The multi-task loss consists of three components:  \\ \\
1. The holistic loss $\mathcal{L}$.

\begin{equation}
    \mathcal{L} = -\sum_{i=1}^{81}y^i\log{p^i}
\end{equation}
2. The first digit loss $\mathcal{L}_1$.
\begin{equation}
    \mathcal{L}_1 = -\sum_{j=1}^{11}y_1^j\log{p_1^j}
\end{equation}
3. The  second digit loss $\mathcal{L}_2$.
 \begin{equation}
    \mathcal{L}_2 = -\sum_{k=1}^{11}y_2^k\log{p_2^k}
\end{equation}
 Each of the three losses is a cross-entropy loss between the ground truth and the predicted distribution.
The overall loss $\mathcal{L}_{tot}$ is given by

\begin{equation}
     \mathcal{L}_{tot} = \alpha*\mathcal{L} + \beta*\mathcal{L}_1+ \gamma*\mathcal{L}_2 
\end{equation}
   where $\alpha, \beta, \gamma$ denote weights given to each loss such that $\alpha + \beta + \gamma = 1$. Also, 
   \begin{equation}
       \beta*\mathcal{L}_1+ \gamma*\mathcal{L}_2
   \end{equation} 
   is the overall digit-wise loss and  $\beta + \gamma$ is the total weight given to the digit-wise loss.

\subsection{Training details}
For data augmentation, we perform color jittering with high values of the $hue$ parameter. Affine transformation are however not performed since they led to a decrease in performance. This is because transformations such as scaling can often make a jersey number not visible since each image has a different scale. The training is done for $10,000$ iterations  with Adam optimizer initial learning rate of $.001$ and $L2$ weight decay of $.001$. The learning rate is decreased by a factor of $0.33$ after $2000,4000,6000$ and $7000$ iterations.  A batch size 100 is used on a single 1080Ti GPU.

\begin{table*}[!t]

    \centering
    \caption{ Comparison of accuracy values with different values of loss weight coefficients for the multi-task setting }
    \footnotesize
    \setlength{\tabcolsep}{0.2cm}
    \begin{tabular}{c|c|c|c|c|c|c}\hline
  
    $\alpha$ & $\beta$ & $\gamma$   & Test Acc & Precision & Recall & F1 score \\\hline
      $1$ & $0$ & $0$         &  $87.6$  & $90.9$ & $87.7$ & $88.7$  \\
     $0.8$ & $0.1$ & $0.1$        &  $87.8$  & $92.0$ & $87.3$ & $89.0$\\
       $0.5$ & $0.25$ & $0.25$ & $89.1$& $92.3$ & $89.1$ & $90.2$ \\
     $0.33$ & $0.33$ & $0.33$ &  $88.4$ & $92.7$ & $88.4$ & $90.0$\\
      $0.3$ & $0.35$ & $0.35$ &  $\textbf{89.6}$ & $\textbf{93.6}$ & $\textbf{89.6}$ & $\textbf{91.2}$\\
       $0.2$ & $0.4$ & $0.4$ &  $89.6$ & $92.8$ & $89.6$ & $90.9$\\
        $0.1$ & $0.45$ & $0.45$ &  $89.0$ & $92.9$ & $89.07$ & $90.6$\\
         $0$ & $0.5$ & $0.5$  & $88.1$ & $92.5$ & $88.1$  & $89.9$\\
    \end{tabular}
    \label{table:ablation}
\end{table*}

\begin{figure*}
	\begin{center}
		\subfloat[Actual:24; Predicted:21]{
		    \includegraphics[width=0.25\linewidth]{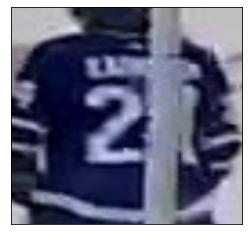}
		}
		\subfloat[Actual:16; Predicted:18]{
		    \includegraphics[width=0.25\linewidth]{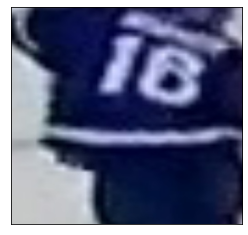}
		}
		\subfloat[Actual:12; Predicted:2]{
		    \includegraphics[width=0.25\linewidth]{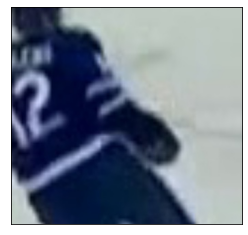}
		}
		\subfloat[Actual:72; Predicted:77]{
		    \includegraphics[width=0.25\linewidth]{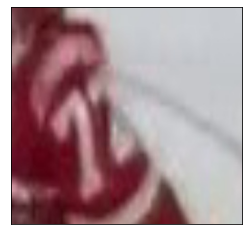}
		}
	\end{center}
	\caption{Some common sources of error are (a) occlusions from external sources, (b) folding of jersey, (c) faulty bounding boxes, and (d) camera viewpoints not covering the whole jersey.}
	\label{fig:images_fc}
\end{figure*}
\section{Results and discussion}
We compare the proposed multi-task loss with holistic and digit-wise losses by simply removing the other loss branch from the network. For the digit-wise setting, a predicted number is classified correctly when both of its digits are classified correctly. From Table \ref{table:acc_values}, the multi-task loss gives an accuracy of $89.6\%$ and a macro averaged F1 score of  $91.2\%$ and outperforms the holistic (accuracy $87.6 \%$ ) and digit-wise losses (accuracy $88.1 \%$ ). Fig. \ref{fig:val_acc} shows the validation accuracy for the three settings during $10,000$ training iterations. The multi-task loss outperforms holistic and digit-wise losses during training.\par 

We  implemented Gerke \textit{et al.} \cite{gerke} model on our dataset and found the performance low ($45.7 \%$ test accuracy). We believe that the reasons for this low performance are (1) The much bigger size of our dataset compared to Gerke \textit{et al.} \cite{gerke} that lowered the generalizability  Gerke \textit{et al.}. (2) Ice hockey is a more challenging domain for jersey number identification than soccer due to high motion blur from fast moving-camera. \par

We also implemented the version of  Li \textit{et. al.} \cite{li} on our dataset without using spatial transformer localization loss since it requires ‘quadrangle’ annotations as mentioned in   Li \textit{et. al.} \cite{li}. The accuracy obtained was $80.0\%$ with F1 score of $82.5\%$ (Table \ref{table:acc_values}). We further replaced the classification cross entropy loss function in Li \textit{et al.} \cite{li} with the proposed loss function and found an improvement in accuracy of $1.6\%$ ($81.6\%$ accuracy) and F1 score of $1.2\%$ ($83.7\%$ F1 score) demonstrating the effectiveness of the proposed loss function.  We could not compare our model with Liu \textit{et al} \cite{liu} since training Liu \textit{et al} \cite{liu} model requires digit level bounding boxes and human keypoint annotations which our dataset does not have and there are no trained models provided by the authors to be used publicly for testing. \par

\begin{table}[!t]

    \centering
    \caption{Comparison of datasets in literature}
    \footnotesize
    \setlength{\tabcolsep}{0.2cm}
    \begin{tabular}{c|c}\hline
  
       Dataset & Number of images \\\hline
     Gerke \textit{et al.} \cite{gerke}        &  $8,281$  \\
       Liu \textit{et al.} \cite{liu}  & $3,567$  \\
      Ours &  $\textbf{54,251}$\\
   
    \end{tabular}
    \label{table:dataset_comp}
\end{table}


Fig. \ref{fig:images_fc} shows some interesting failure cases. Partial occlusions are common and can result in misinterpretation of jersey numbers (Fig. \ref{fig:images_fc} part a). Other sources of failures are folding of jersey leading to errors Fig. \ref{fig:images_fc} part b, jersey numbers not fully present in player bounding boxes  (Fig. \ref{fig:images_fc} part c) and jersey number occluded due to camera viewpoints  (Fig. \ref{fig:images_fc} part d).

\begin{table}[!t]

    \centering
    \caption{Comparison of accuracy values with different backbone networks}
    \footnotesize
    \setlength{\tabcolsep}{0.2cm}
    \begin{tabular}{c|c|c|c|c}\hline
  
       Backbone & Test Acc & Precision & Recall & F1 score \\\hline
     Mobilenetv2        &  $87.9$  & $91.8$ & $87.9$ & $89.3$\\
       Resnet18  & $89.1$ & $92.5$ & $89.1$  & $90.3$\\
      Resnet34 &  $\textbf{89.6}$ & $\textbf{93.6}$ & $\textbf{89.6}$ & $\textbf{91.2}$\\
   
    \end{tabular}
    \label{table:backbone}
\end{table}

\subsection{Ablation study}
We perform an ablation study on the loss weights $\alpha, \beta$, and~$\gamma$ to determine how the digit-wise and holistic losses affect accuracy. The analysis can be seen in Table \ref{table:ablation}. We observe that giving a higher weight to the digit-wise loss ($\beta + \gamma = 0.7$) gives the highest accuracy  ($89.6 \%$) and F1 score ($91.2 \%$). However, having a high value of holistic loss weight ($\alpha = 0.8$) results in a lower accuracy($87.8 \%$) and F1 score ($89.0\%$).  This makes sense because on its own, the digit-wise loss gives better accuracy compared to holistic loss ( Table \ref{table:acc_values}). However, as $\beta+\gamma$ is further increased to $0.9$ the accuracy decreases ($89\%$). This demonstrates that holistic and digit-wise losses complement each other when an appropriate weight is given to  both losses. The accuracy is maximized when the digit-wise loss is given slightly more than double the weight of the holistic loss. The best values are $\alpha=0.3, \beta = 0.35$ and $\gamma=0.35$. \\

\begin{table}[!t]

    \centering
    \caption{Comparison of accuracy values with holistic, digit-wise and multi-task settings }
    \footnotesize
    \setlength{\tabcolsep}{0.2cm}
    \begin{tabular}{c|c|c|c|c}\hline
  
       Method & Test Acc & Precision & Recall & F1 score \\\hline
     Holistic        &  $87.6$  & $90.9$ & $87.7$ & $88.7$\\
       digit-wise  & $88.1$ & $92.5$ & $88.1$  & $89.9$\\
      multi-task &  $\textbf{89.6}$ & $\textbf{93.6}$ & $\textbf{89.6}$ & $\textbf{91.2}$\\
   
      Li \textit{et al.} \cite{li} & $80.0$ & $87.1$ & $80.0$ &  $82.5$\\
      Li \textit{et al.} \cite{li}(proposed loss) & $81.6$ & $87.9$ & $81.6$ &  $83.7$\\
    Gerke \textit{et al.} \cite{gerke} & $45.7$ &$58.5$ &$45.7$ & $48.2$\\
    \end{tabular}
    \label{table:acc_values}
\end{table}

We also do an ablation study on the backbone network used in the experiment in Table \ref{table:backbone}. Two additional  backbones were tested: Resnet18 \cite{resnet}, Mobilenetv2 \cite{mobilenet}, while keeping  other parameters including the loss weights $\alpha, \beta, \gamma$ fixed to their optimal values of $0.3, 0.35, 0.35$. Resnet 34 showed the best performace followed by Resnet18 and Mobilenetv2. We did not test bigger networks such as  Resnet 50 since it could not fit a batch size of 100 on a single GPU.

\section{Conclusion}
In this paper, we introduce a simple multi-task learning network for player's jersey number recognition in ice hockey broadcast video frames. We also create a new dataset with more than $50,000$ images to test the network. The network learns both the holistic and digit-wise representations of jersey number labels which resulted in improved regularization and accuracy.  The methodology is however, task agnostic and can be used in other number recognition tasks.

\section{Acknowledgement}
This work was supported by Stathletes through the Mitacs Accelerate Program and Natural Sciences and Engineering Research Council of Canada (NSERC).

\bibliographystyle{ACM-Reference-Format}
\bibliography{sample-base}


\begin{thebibliography}{14}


\ifx \showCODEN    \undefined \def \showCODEN     #1{\unskip}     \fi
\ifx \showDOI      \undefined \def \showDOI       #1{#1}\fi
\ifx \showISBNx    \undefined \def \showISBNx     #1{\unskip}     \fi
\ifx \showISBNxiii \undefined \def \showISBNxiii  #1{\unskip}     \fi
\ifx \showISSN     \undefined \def \showISSN      #1{\unskip}     \fi
\ifx \showLCCN     \undefined \def \showLCCN      #1{\unskip}     \fi
\ifx \shownote     \undefined \def \shownote      #1{#1}          \fi
\ifx \showarticletitle \undefined \def \showarticletitle #1{#1}   \fi
\ifx \showURL      \undefined \def \showURL       {\relax}        \fi
\providecommand\bibfield[2]{#2}
\providecommand\bibinfo[2]{#2}
\providecommand\natexlab[1]{#1}
\providecommand\showeprint[2][]{arXiv:#2}

\bibitem[\protect\citeauthoryear{Gerke, Linnemann, and Müller}{Gerke
  et~al\mbox{.}}{2017}]%
        {GERKE2017105}
\bibfield{author}{\bibinfo{person}{Sebastian Gerke}, \bibinfo{person}{Antje
  Linnemann}, {and} \bibinfo{person}{Karsten Müller}.}
  \bibinfo{year}{2017}\natexlab{}.
\newblock \showarticletitle{Soccer player recognition using spatial
  constellation features and jersey number recognition}.
\newblock \bibinfo{journal}{\emph{Computer Vision and Image Understanding}}
  \bibinfo{volume}{159} (\bibinfo{year}{2017}), \bibinfo{pages}{105 -- 115}.
\newblock
\showISSN{1077-3142}
\urldef\tempurl%
\url{https://doi.org/10.1016/j.cviu.2017.04.010}
\showDOI{\tempurl}
\newblock
\shownote{Computer Vision in Sports.}


\bibitem[\protect\citeauthoryear{{Gerke}, {Müller}, and {Schäfer}}{{Gerke}
  et~al\mbox{.}}{2015}]%
        {gerke}
\bibfield{author}{\bibinfo{person}{S. {Gerke}}, \bibinfo{person}{K. {Müller}},
  {and} \bibinfo{person}{R. {Schäfer}}.} \bibinfo{year}{2015}\natexlab{}.
\newblock \showarticletitle{Soccer Jersey Number Recognition Using
  Convolutional Neural Networks}. In \bibinfo{booktitle}{\emph{2015 IEEE
  International Conference on Computer Vision Workshop (ICCVW)}}.
  \bibinfo{pages}{734--741}.
\newblock
\urldef\tempurl%
\url{https://doi.org/10.1109/ICCVW.2015.100}
\showDOI{\tempurl}


\bibitem[\protect\citeauthoryear{Goodfellow, Bulatov, Ibarz, Arnoud, and
  Shet}{Goodfellow et~al\mbox{.}}{2014}]%
        {goodfellow}
\bibfield{author}{\bibinfo{person}{Ian Goodfellow}, \bibinfo{person}{Yaroslav
  Bulatov}, \bibinfo{person}{Julian Ibarz}, \bibinfo{person}{Sacha Arnoud},
  {and} \bibinfo{person}{Vinay Shet}.} \bibinfo{year}{2014}\natexlab{}.
\newblock \showarticletitle{Multi-digit Number Recognition from Street View
  Imagery using Deep Convolutional Neural Networks}. In
  \bibinfo{booktitle}{\emph{ICLR2014}}.
\newblock


\bibitem[\protect\citeauthoryear{{He}, {Zhang}, {Ren}, and {Sun}}{{He}
  et~al\mbox{.}}{2016}]%
        {resnet}
\bibfield{author}{\bibinfo{person}{K. {He}}, \bibinfo{person}{X. {Zhang}},
  \bibinfo{person}{S. {Ren}}, {and} \bibinfo{person}{J. {Sun}}.}
  \bibinfo{year}{2016}\natexlab{}.
\newblock \showarticletitle{Deep Residual Learning for Image Recognition}. In
  \bibinfo{booktitle}{\emph{2016 IEEE Conference on Computer Vision and Pattern
  Recognition (CVPR)}}. \bibinfo{pages}{770--778}.
\newblock
\urldef\tempurl%
\url{https://doi.org/10.1109/CVPR.2016.90}
\showDOI{\tempurl}


\bibitem[\protect\citeauthoryear{{Hou}, {Zeng}, {Cai}, {Zhu}, {Cao}, and
  {Hou}}{{Hou} et~al\mbox{.}}{2017}]%
        {hou}
\bibfield{author}{\bibinfo{person}{J. {Hou}}, \bibinfo{person}{H. {Zeng}},
  \bibinfo{person}{L. {Cai}}, \bibinfo{person}{J. {Zhu}}, \bibinfo{person}{J.
  {Cao}}, {and} \bibinfo{person}{J. {Hou}}.} \bibinfo{year}{2017}\natexlab{}.
\newblock \showarticletitle{Handwritten numeral recognition using multi-task
  learning}. In \bibinfo{booktitle}{\emph{2017 International Symposium on
  Intelligent Signal Processing and Communication Systems (ISPACS)}}.
  \bibinfo{pages}{155--158}.
\newblock
\urldef\tempurl%
\url{https://doi.org/10.1109/ISPACS.2017.8266464}
\showDOI{\tempurl}


\bibitem[\protect\citeauthoryear{Kim, Park, Oh, and Kang}{Kim
  et~al\mbox{.}}{2017}]%
        {kim}
\bibfield{author}{\bibinfo{person}{Hong-Hyun Kim}, \bibinfo{person}{Je-Kang
  Park}, \bibinfo{person}{Joo-Hee Oh}, {and} \bibinfo{person}{Dong Kang}.}
  \bibinfo{year}{2017}\natexlab{}.
\newblock \showarticletitle{Multi-task convolutional neural network system for
  license plate recognition}.
\newblock \bibinfo{journal}{\emph{International Journal of Control, Automation
  and Systems}}  \bibinfo{volume}{15} (\bibinfo{date}{12}
  \bibinfo{year}{2017}), \bibinfo{pages}{2942--2949}.
\newblock
\urldef\tempurl%
\url{https://doi.org/10.1007/s12555-016-0332-z}
\showDOI{\tempurl}


\bibitem[\protect\citeauthoryear{{Li}, {Xu}, {Liu}, {Li}, and {Wang}}{{Li}
  et~al\mbox{.}}{2018}]%
        {li}
\bibfield{author}{\bibinfo{person}{G. {Li}}, \bibinfo{person}{S. {Xu}},
  \bibinfo{person}{X. {Liu}}, \bibinfo{person}{L. {Li}}, {and}
  \bibinfo{person}{C. {Wang}}.} \bibinfo{year}{2018}\natexlab{}.
\newblock \showarticletitle{Jersey Number Recognition with Semi-Supervised
  Spatial Transformer Network}. In \bibinfo{booktitle}{\emph{2018 IEEE/CVF
  Conference on Computer Vision and Pattern Recognition Workshops (CVPRW)}}.
  \bibinfo{pages}{1864--18647}.
\newblock
\urldef\tempurl%
\url{https://doi.org/10.1109/CVPRW.2018.00231}
\showDOI{\tempurl}


\bibitem[\protect\citeauthoryear{{Liu} and {Bhanu}}{{Liu} and {Bhanu}}{2019}]%
        {liu}
\bibfield{author}{\bibinfo{person}{H. {Liu}} {and} \bibinfo{person}{B.
  {Bhanu}}.} \bibinfo{year}{2019}\natexlab{}.
\newblock \showarticletitle{Pose-Guided R-CNN for Jersey Number Recognition in
  Sports}. In \bibinfo{booktitle}{\emph{2019 IEEE/CVF Conference on Computer
  Vision and Pattern Recognition Workshops (CVPRW)}}.
  \bibinfo{pages}{2457--2466}.
\newblock
\urldef\tempurl%
\url{https://doi.org/10.1109/CVPRW.2019.00301}
\showDOI{\tempurl}


\bibitem[\protect\citeauthoryear{Ren, He, Girshick, and Sun}{Ren
  et~al\mbox{.}}{2015}]%
        {fasterrcnn}
\bibfield{author}{\bibinfo{person}{Shaoqing Ren}, \bibinfo{person}{Kaiming He},
  \bibinfo{person}{Ross Girshick}, {and} \bibinfo{person}{Jian Sun}.}
  \bibinfo{year}{2015}\natexlab{}.
\newblock \showarticletitle{Faster R-CNN: Towards Real-Time Object Detection
  with Region Proposal Networks}. In \bibinfo{booktitle}{\emph{Proceedings of
  the 28th International Conference on Neural Information Processing Systems -
  Volume 1}} (Montreal, Canada) \emph{(\bibinfo{series}{NIPS'15})}.
  \bibinfo{publisher}{MIT Press}, \bibinfo{address}{Cambridge, MA, USA},
  \bibinfo{pages}{91–99}.
\newblock


\bibitem[\protect\citeauthoryear{Ruder}{Ruder}{2017}]%
        {ruder}
\bibfield{author}{\bibinfo{person}{Sebastian Ruder}.}
  \bibinfo{year}{2017}\natexlab{}.
\newblock \showarticletitle{An Overview of Multi-Task Learning in Deep Neural
  Networks}.
\newblock \bibinfo{journal}{\emph{CoRR}}  \bibinfo{volume}{abs/1706.05098}
  (\bibinfo{year}{2017}).
\newblock
\showeprint[arxiv]{1706.05098}
\urldef\tempurl%
\url{http://arxiv.org/abs/1706.05098}
\showURL{%
\tempurl}


\bibitem[\protect\citeauthoryear{{Sandler}, {Howard}, {Zhu}, {Zhmoginov}, and
  {Chen}}{{Sandler} et~al\mbox{.}}{2018}]%
        {mobilenet}
\bibfield{author}{\bibinfo{person}{M. {Sandler}}, \bibinfo{person}{A.
  {Howard}}, \bibinfo{person}{M. {Zhu}}, \bibinfo{person}{A. {Zhmoginov}},
  {and} \bibinfo{person}{L. {Chen}}.} \bibinfo{year}{2018}\natexlab{}.
\newblock \showarticletitle{MobileNetV2: Inverted Residuals and Linear
  Bottlenecks}. In \bibinfo{booktitle}{\emph{2018 IEEE/CVF Conference on
  Computer Vision and Pattern Recognition}}. \bibinfo{pages}{4510--4520}.
\newblock
\urldef\tempurl%
\url{https://doi.org/10.1109/CVPR.2018.00474}
\showDOI{\tempurl}


\bibitem[\protect\citeauthoryear{Senocak, Oh, Kim, and So~Kweon}{Senocak
  et~al\mbox{.}}{2018}]%
        {Senocak_2018_CVPR_Workshops}
\bibfield{author}{\bibinfo{person}{Arda Senocak}, \bibinfo{person}{Tae-Hyun
  Oh}, \bibinfo{person}{Junsik Kim}, {and} \bibinfo{person}{In So~Kweon}.}
  \bibinfo{year}{2018}\natexlab{}.
\newblock \showarticletitle{Part-Based Player Identification Using Deep
  Convolutional Representation and Multi-Scale Pooling}. In
  \bibinfo{booktitle}{\emph{Proceedings of the IEEE Conference on Computer
  Vision and Pattern Recognition (CVPR) Workshops}}.
\newblock


\bibitem[\protect\citeauthoryear{Ye, Jiang, Liu, and Gao}{Ye
  et~al\mbox{.}}{2005}]%
        {hc}
\bibfield{author}{\bibinfo{person}{Qixiang Ye}, \bibinfo{person}{Shuqiang
  Jiang}, \bibinfo{person}{Yang Liu}, {and} \bibinfo{person}{Wen Gao}.}
  \bibinfo{year}{2005}\natexlab{}.
\newblock \showarticletitle{Jersey number detection in sports video for athlete
  identification}.
\newblock \bibinfo{journal}{\emph{Proc SPIE}}  \bibinfo{volume}{5960}
  (\bibinfo{date}{07} \bibinfo{year}{2005}), \bibinfo{pages}{1599--1606}.
\newblock
\urldef\tempurl%
\url{https://doi.org/10.1117/12.632735}
\showDOI{\tempurl}


\bibitem[\protect\citeauthoryear{Šaric, Dujmic, Papic, and Rožic}{Šaric
  et~al\mbox{.}}{2008}]%
        {Saric2008PlayerNL}
\bibfield{author}{\bibinfo{person}{Matko Šaric}, \bibinfo{person}{Hrvoje
  Dujmic}, \bibinfo{person}{Vladan Papic}, {and} \bibinfo{person}{Nikola
  Rožic}.} \bibinfo{year}{2008}\natexlab{}.
\newblock \showarticletitle{Player Number Localization and Recognition in
  Soccer Video using HSV Color Space and Internal Contours}.
\newblock \bibinfo{journal}{\emph{International Journal of Electrical and
  Computer Engineering}} \bibinfo{volume}{2}, \bibinfo{number}{7}
  (\bibinfo{year}{2008}), \bibinfo{pages}{1408 -- 1412}.
\newblock
\showISSN{eISSN: 1307-6892}
\urldef\tempurl%
\url{https://publications.waset.org/vol/19}
\showURL{%
\tempurl}


\end{thebibliography}

\end{document}